\documentclass[11pt]{article}

\usepackage[margin=1in,top=1.25in]{geometry}

\usepackage[T1]{fontenc}
\usepackage[utf8]{inputenc}
\usepackage{times}
\usepackage{microtype}

\usepackage{amsmath,amssymb}

\usepackage{booktabs}
\usepackage{multirow}
\usepackage{caption}

\usepackage{graphicx}
\graphicspath{{../figures/}{../appendix/}}
\usepackage{xcolor}
\usepackage{placeins}
\usepackage{stfloats}
\usepackage{float}   

\usepackage{natbib}
\usepackage[colorlinks=true,linkcolor=blue,citecolor=blue,urlcolor=blue,
            pdftitle={Linguistic Accommodation Between Neurodivergent Communities on Reddit},
            pdfauthor={Saad Mankarious, Iyad Ait Hou, Nour Zein, Aya Zirikly}]{hyperref}

\usepackage{authblk}

\makeatletter
\def\@maketitle{%
  \newpage\null
  \noindent\rule{\linewidth}{1.5pt}
  \vskip 0.6em
  {\Large\bfseries\centering\@title\par}
  \vskip 0.5em
  \noindent\rule{\linewidth}{1.5pt}
  \vskip 0.5em
  {\centering\@author\par}
  \vskip 0.8em
}
\makeatother

\title{Mirroring Minds: Asymmetric Linguistic Accommodation and Diagnostic
Identity in ADHD and Autism Reddit Communities}

\author[1]{Saad Mankarious\thanks{saadm@gwu.edu (Corresponding Author)}}
\author[2]{Nour Zeid\thanks{nourzzeid@gmail.com (Corresponding Author)}}
\author[1]{Iyad Ait Hou\thanks{iyad.aithou@gwu.edu}}
\author[1]{Rebecca Hwa\thanks{rebecca.hwa@gwu.edu}}
\author[1]{Aya Zirikly\thanks{ayah.zirikly@gwu.edu}}

\affil[1]{The George Washington University}
\affil[2]{Texas A\&M University}

\date{}

\begin{document}
\maketitle
\begin{abstract}
Social media research on mental health has focused predominantly on detecting and diagnosing conditions at the individual level. In this work, we shift attention to \emph{intergroup} behavior, examining how two prominent neurodivergent communities, ADHD and autism, adjust their language when engaging with each other on Reddit. Grounded in Communication Accommodation Theory (CAT), we first establish that each community maintains a distinct linguistic profile as measured by Language Inquiry and Word Count Lexicon (LIWC). We then show that these profiles shift in opposite directions when users cross community boundaries: features that are elevated in one group's home community decrease when its members post in the other group's space, and vice versa, consistent with convergent accommodation. The involvement of topic-independent summary variables (Authentic, Clout) in these shifts provides partial evidence against a purely topical explanation. Finally, in an exploratory longitudinal analysis around the moment of public diagnosis disclosure, we find that its effects on linguistic style are small and, in some cases, directionally opposite to cross-community accommodation, providing initial evidence that situational audience adaptation and longer-term identity processes may involve different mechanisms. Our findings contribute to understanding intergroup communication dynamics among neurodivergent populations online and carry implications for community moderation and clinical perspectives on these conditions.
\end{abstract}

\section{Introduction}

Language carries diagnostic identity. When a person diagnosed with ADHD writes on Reddit, their posts bear measurable traces of that identity,more temporal references, more achievement-oriented language, more markers of raw authenticity. A person diagnosed with autism leaves a different trace: more social references, more third-person framing, higher linguistic clout. These are not stereotypes; they are statistical regularities that emerge robustly across thousands of users and hundreds of thousands of posts. But what happens to these identity-linked markers when a user steps outside their home community and enters someone else's?

\newpage

\begin{figure}[H]
    \centering
    \includegraphics[width=0.7\textwidth]{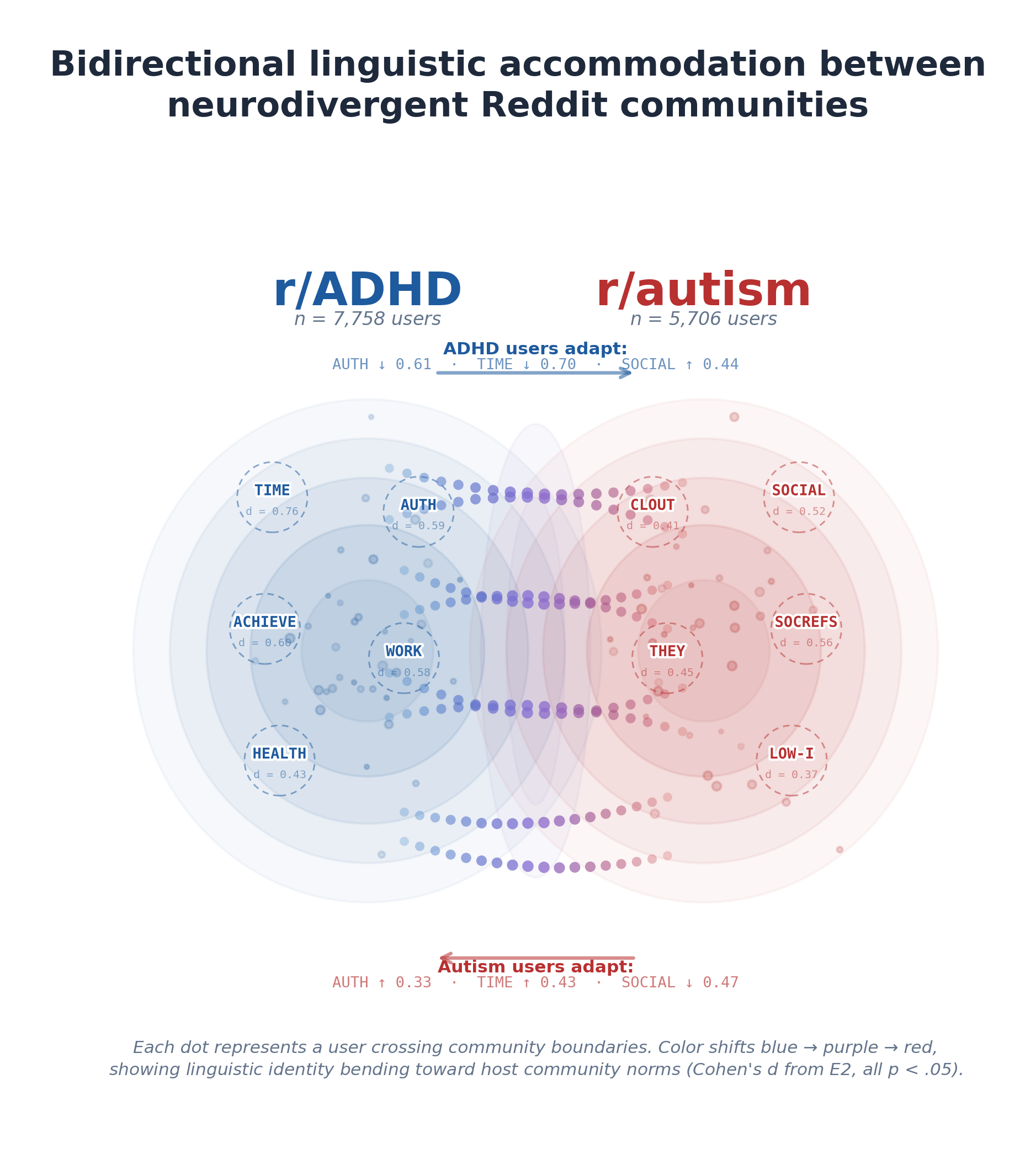}
    \caption{Bidirectional linguistic accommodation between neurodivergent Reddit
    communities. r/ADHD (blue, $n = 7{,}758$) is characterised by higher
    \textsc{time}, \textsc{Authentic}, \textsc{achieve}, and \textsc{work};
    r/autism (red, $n = 5{,}706$) by higher \textsc{Social}, \textsc{socrefs},
    \textsc{Clout}, and \textsc{they}. Each dot is a cross-posting user; colour
    shifts (blue $\to$ purple $\to$ red) indicate accommodation toward host norms.}
    \label{fig:communities}
\end{figure}

\noindent

This question sits at the intersection of two problems that have largely been studied in isolation. The first is the computational analysis of mental health in social media, which has produced a rich literature on detecting and characterizing conditions at the individual level \citep{chancellor2020} but has overwhelmingly treated users as isolated data points rather than members of dynamic communities. The second is Communication Accommodation Theory \citep[CAT;][]{giles2015}, which has a decades-long tradition of studying how people adjust their communicative style in response to social context,converging toward, diverging from, or maintaining their styles depending on audience and identity goals. CAT has been validated in digital settings \citep{danescu2011,boghrati2018,tamburrini2015}, and extended to affective dimensions by \citet{bernhold2022}, but has rarely been applied to intergroup dynamics in digital mental health communities, where the stakes of linguistic identity are particularly high.

The gap matters for a concrete reason. Online neurodivergent communities are among the largest and most active mental health spaces on the internet. On Reddit alone, r/ADHD and r/autism together host millions of subscribers who use these spaces not just for information exchange but for identity construction, peer support, and diagnostic sense-making \citep{rideout2021}. Members of these communities regularly cross-post: an ADHD-diagnosed user may seek advice in r/autism, or an autism-diagnosed user may share experiences in r/ADHD. When they do, they face an implicit communicative choice,maintain their own group's linguistic style, or adapt to the norms of the host community. The outcome of this choice has practical implications for inclusivity, information flow across diagnostic boundaries, and platform moderation.

In this work, we apply CAT to examine intergroup communication between r/ADHD and r/autism on Reddit, measuring accommodation patterns when users diagnosed with one condition post in the other condition's subreddit. We further analyze how these patterns change around a critical moment in a user's online trajectory: the public disclosure of their diagnosis. Our study is, to our knowledge, the first to examine cross-community linguistic accommodation between distinct neurodivergent groups.

\paragraph{Hypotheses and research question.} Grounded in CAT and the prior work reviewed above, we test two hypotheses and pose one exploratory research question:

\begin{itemize}
    \item[\textbf{H1}] \textbf{(Distinct Baseline Styles)} ADHD and autism communities exhibit measurably distinct linguistic profiles when posting in their respective home communities.
    \item[\textbf{H2}] \textbf{(Cross-Community Convergence)} When users post in the other community, their linguistic style shifts toward the host community's baseline, consistent with convergent accommodation. Specifically, categories where the two groups differ at baseline will show shifts in the direction of the host community's norms.
    \item[\textbf{RQ}] \textbf{(Diagnosis Disclosure)} Does public disclosure of a diagnosis produce within-user linguistic changes, and if so, are these changes consistent with, independent of, or opposed to the situational accommodation captured in H2? We treat this as exploratory given the inherent power limitations of longitudinal within-user analysis.
\end{itemize}

\paragraph{Contributions.} Our work makes the following contributions:
\begin{itemize}
    \item We provide the first study of cross-community linguistic accommodation between neurodivergent groups on social media, shifting the focus from individual-level diagnosis detection to intergroup behavioral dynamics.
    \item We establish distinct linguistic profiles for ADHD and autism communities and demonstrate that these profiles undergo systematic, bidirectional shifts when users cross community boundaries, with medium effect sizes and a mirror-image pattern across multiple LIWC dimensions.
    \item We present an exploratory longitudinal analysis of pre-/post-diagnosis linguistic changes, finding initial evidence that situational accommodation and identity-related shifts may operate through different mechanisms,though the small effect sizes warrant cautious interpretation.
\end{itemize}

\section{Related Work}

\subsection{Communication Accommodation Theory}
Communication Accommodation Theory (CAT) provides a foundational framework for understanding how individuals adjust their communication styles in social interactions \citep{giles1973}. CAT posits that speakers may either converge toward their interlocutor's style to seek approval or diverge to emphasize distinctiveness. The theory has been widely applied across contexts, including psychotherapy, where accommodation patterns reflect therapeutic rapport and power dynamics \citep{ferrara1991}.

\subsection{CAT in Social Media and Online Communities}
Recent work has extended CAT to digital contexts, particularly social media platforms where interaction patterns are publicly observable. \citet{danescu2011} developed a rigorous framework for measuring linguistic accommodation on Twitter, demonstrating that users adapt their language to align with conversational partners. Their approach focused on function words as stylistic markers and introduced probabilistic methods to isolate accommodation from topic-driven similarity. On Reddit, \citet{boghrati2018} showed that commenters systematically accommodate to the syntactic style of original posts, providing evidence for accommodation across diverse interactional contexts. \citet{tamburrini2015} found that members of online communities converge in their linguistic practices over time, and \citet{kovacs2021} showed that affective convergence follows a similar pattern within communities. Of direct relevance to our work, \citet{sharma2018} examined accommodation across 55 Reddit mental health communities, finding a positive relationship between the degree of accommodation exhibited in posts and the level of social support received. However, most of this work has examined accommodation \emph{within} a single community or between pairs of interlocutors. Our work differs by examining accommodation \emph{across} two distinct communities,specifically, what happens when a member of one diagnostic community posts in the other's space.

\subsection{LIWC and Mental Health Discourse}
Linguistic Inquiry and Word Count (LIWC) has been widely employed to characterize mental health discourse in online communities \citep{pennebaker2015}. \citet{lyons2018} analyzed language patterns across multiple mental health conditions on discussion forums, identifying distinctive linguistic markers associated with various conditions. Within autism research, LIWC has been used to identify subtle language differences in narrative production \citep{boorse2019}. ADHD communities have been less studied linguistically, though work on ADHD self-disclosure suggests that these users employ distinctive patterns around achievement, temporal orientation, and self-reference \citep{guntuku2019}. A growing body of work uses NLP to detect mental health conditions from social media text \citep{chancellor2020}, but this line of research treats users as isolated individuals rather than members of interacting communities. Our work extends this by comparing ADHD and autism communities directly, both in baseline styles and accommodation patterns.

\subsection{Identity, Disclosure, and Mental Health on Reddit}
The moment at which an individual publicly discloses a mental health diagnosis has been recognized as a pivotal identity event in both clinical and social-psychological literature \citep{corrigan2015}. Reddit has emerged as a valuable source for studying mental health discourse due to its anonymous, community-based structure \citep{dechoudhury2014}. On Reddit, self-disclosure of a diagnosis often functions as an identity claim that restructures how an individual participates in community discourse \citep{dearmond2020}. More recently, researchers have examined identity formation and disclosure processes in neurodivergent populations, highlighting the developmental trajectory of diagnostic identity \citep{depape2025} and how autism social identification predicts disclosure behaviors \citep{togher2023}. Prior work has examined how disclosure affects reception by others, but less attention has been given to how it changes the discloser's own linguistic behavior over time. Our longitudinal analysis addresses this gap.

\section{Data}

\begin{table*}[!ht]
\centering
\small
\begin{tabular}{lrrrrrrr}
\toprule
\textbf{Dataset} & \textbf{Users} & \textbf{Posts} & \textbf{Avg Words} & \textbf{Std} & \textbf{Subreddit} & \textbf{Comments} & \textbf{Submissions} \\
\midrule
ADHD in ADHD     & 7{,}758 & 199{,}020 & 85.94 & 112.21 & r/ADHD   & 171{,}028 & 27{,}992 \\
ADHD in Autism   & 788  & 12{,}621  & 70.96 & 113.82 & r/autism & 11{,}245 & 1{,}376 \\
Autism in Autism & 5{,}706 & 276{,}159 & 64.93 & 98.54  & r/autism & 248{,}597 & 27{,}562 \\
Autism in ADHD   & 2{,}305 & 38{,}470  & 84.01 & 108.22 & r/ADHD   & 31{,}947  & 6{,}523 \\
\bottomrule
\end{tabular}
\caption{Descriptive statistics for cross-posting datasets. Users are non-overlapping between ADHD and autism diagnostic groups.}
\label{tab:data_stats}
\end{table*}

\subsection{Source Dataset}
We draw on the Mindset dataset \citep{mindset}, a recently released Reddit corpus in which users are identified via high-precision self-reported diagnosis extraction. Mindset implemented rigorous filtering to ensure label quality, including exclusion of negated diagnoses (e.g., ``I don't have ADHD'') and retention of only confident diagnostic statements. From Mindset, we selected the ADHD, autism, and control subsets. Because our analysis requires cross-community posting, we did not use Mindset's preprocessed posts directly,those are heavily filtered to remove mental health content. Instead, we re-fetched each user's full posting history from Reddit using the same API, retaining posts that include mental health discussions necessary for the accommodation analysis. We additionally collected metadata including interaction scores and parent post identifiers.

\subsection{Cross-Community Partitioning}
\label{sec:cross_partition}
To operationalize accommodation, we partitioned the data into four subsets based on the user's diagnostic group and the subreddit in which they posted (Table~\ref{tab:data_stats}): \emph{ADHD in ADHD} (posts by ADHD-diagnosed users in r/ADHD), \emph{ADHD in Autism} (posts by ADHD-diagnosed users in r/autism), and the symmetric pair for autism-diagnosed users. This subreddit-based partitioning is motivated by two considerations: (1) the vast majority of each group's contributions are concentrated in these two subreddits, and (2) keyword-based topic filtering would be unreliable, since discussions in either subreddit frequently touch on the other condition. We verified that no users appear in both diagnostic groups, ensuring that the four subsets represent cleanly separated populations.

\subsection{Preprocessing}
We excluded posts that had been removed by moderators or that contained fewer than five words after tokenization, as these are unlikely to carry sufficient stylistic signal for LIWC analysis.

\begin{figure*}[t]
    \centering
    \includegraphics[width=0.75\textwidth]{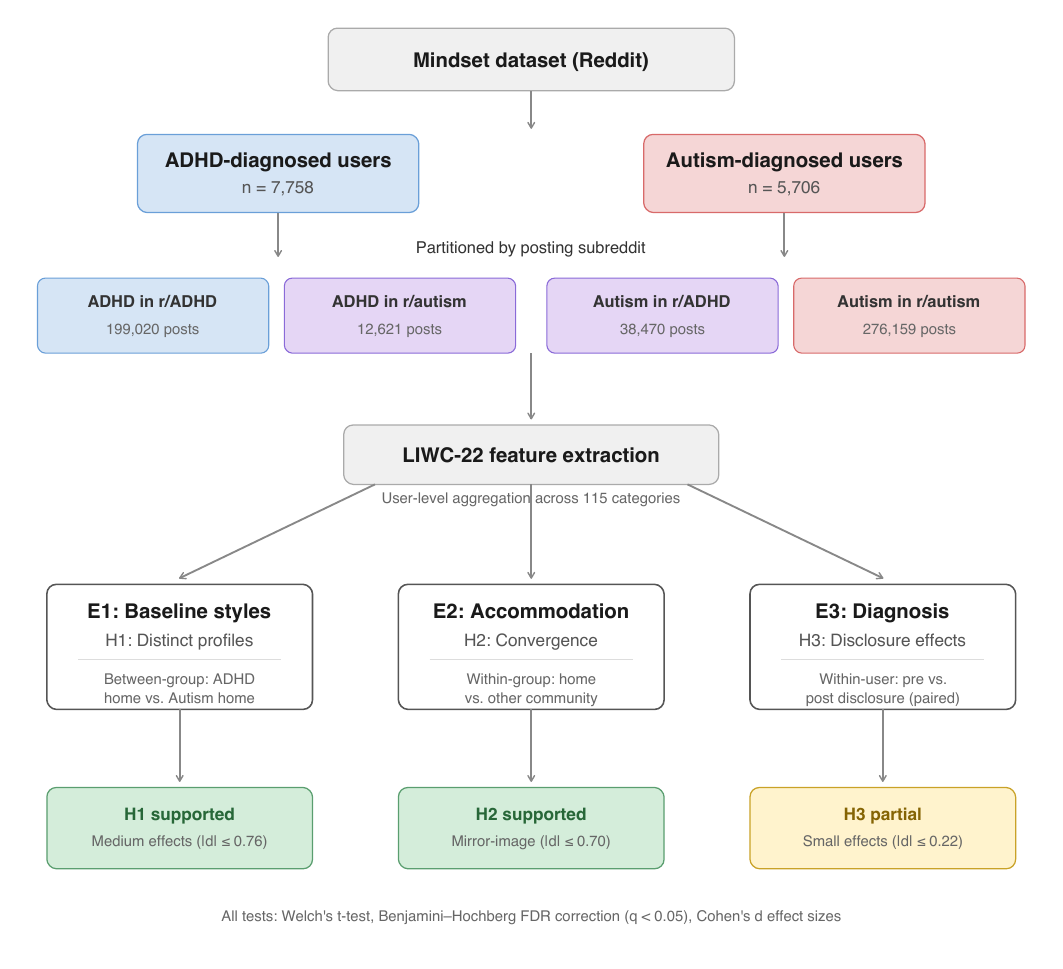}
    \caption{Study design. From the Mindset dataset, we extract ADHD- and autism-diagnosed users, partition their posts by subreddit into four cross-community datasets, apply LIWC-22 feature extraction with user-level aggregation, and conduct three experiments testing baseline differences (E1), cross-community accommodation (E2), and exploratory post-diagnosis changes (E3). Green boxes indicate supported hypotheses; yellow indicates partial/exploratory support.}
    \label{fig:study_design}
\end{figure*}

\section{Experiments}
Figure~\ref{fig:study_design} provides an overview of the experimental pipeline.

\subsection{Linguistic Feature Extraction}
We processed all posts through LIWC-22 \citep{boyd2022}, extracting scores across 115 categories spanning psychological processes, linguistic dimensions, and summary variables. To account for pseudoreplication,the inflation of statistical significance when multiple posts from the same user are treated as independent observations,we aggregated all LIWC scores at the \emph{user} level by averaging across each user's posts within each condition \citep{walls2006}. This ensures that each user contributes a single data point per analysis, yielding conservative but reliable estimates.

\subsection{Statistical Methodology}
\label{sec:stats}
All group comparisons use Welch's $t$-test (unequal variances) with Cohen's $d$ as the effect size measure, accompanied by 95\% confidence intervals computed via the noncentral $t$-distribution (see Appendix~\ref{sec:appendix_ci} for details). Following \citet{benjamini1995}, we apply the Benjamini--Hochberg procedure to control the false discovery rate (FDR) at $q = 0.05$ within each experiment, correcting for the 115 simultaneous tests across LIWC categories. Only categories that remain significant after FDR correction are reported. We adopt standard effect size thresholds: $|d| < 0.20$ negligible, $0.20 \leq |d| < 0.50$ small, $0.50 \leq |d| < 0.80$ medium, $|d| \geq 0.80$ large \citep{cohen1988}.

\subsection{Operationalizing Accommodation}
\label{sec:operationalize}
Classical CAT measures accommodation at the conversational level: speaker A adjusts toward speaker B within a specific exchange. In asynchronous online communities, direct pairing is not always possible. Following \citet{danescu2011} and \citet{tamburrini2015}, we measure \emph{community-level convergence}: whether the same user's linguistic profile shifts toward the host community's aggregate norms when posting in a different subreddit. This within-user design means each user serves as their own control, ruling out subpopulation confounds. The critical evidence for \emph{accommodation} rather than mere \emph{change} is directionality: ADHD users decrease on ADHD-high features and increase on autism-high features when visiting r/autism, and autism users show the exact mirror pattern. Random variation or topic effects would not produce this symmetric, baseline-aligned pattern across multiple independent LIWC categories.

\subsection{Experiment 1: Baseline Linguistic Styles (H1)}

To test H1, we compared diagnosed ADHD users posting in their home community (r/ADHD, $n = 7{,}758$ users) against diagnosed autism users posting in their home community (r/autism, $n = 5{,}706$ users). This between-group comparison characterizes the distinct baseline stylistic profiles against which accommodation effects can be measured.

\subsection{Experiment 2: Cross-Community Accommodation (H2)}

To test H2, we conducted within-group comparisons of each diagnostic group's language when posting in their home community versus the other community:
\begin{itemize}
    \item ADHD users: language in r/ADHD (in-group) vs.\ r/autism (out-group)
    \item Autism users: language in r/autism (in-group) vs.\ r/ADHD (out-group)
\end{itemize}
If convergent accommodation occurs, we expect shifts in the direction of the host community's baseline for categories identified in E1. We additionally examine whether the two groups' shifts are \emph{mirror images},that is, whether categories that shift upward for one group shift downward for the other.

\subsection{Experiment 3: Exploratory Longitudinal Analysis (RQ)}

To address the exploratory research question, we conducted within-user paired comparisons of language use before and after each user's first public diagnosis disclosure. This analysis is restricted to users who have posts in both pre- and post-disclosure periods within a given cross-community dataset ($\geq 3$ posts in each period). The temporal structure of user activity varies across datasets; see Appendix~\ref{sec:appendix_longitudinal} (Table~\ref{tab:timeline}) for average pre- and post-diagnosis posting spans. By comparing E3 effect sizes against E2 effect sizes for the same categories, we assess whether diagnosis disclosure produces changes that are consistent with, independent of, or opposed to situational accommodation. We emphasize that this analysis is exploratory: the within-user design reduces sample sizes substantially, and the resulting effect sizes should be interpreted with appropriate caution.

\section{Results}

\subsection{E1: Baseline Stylistic Differences (H1)}

Table~\ref{tab:baseline_styles_user} presents the 10 LIWC categories with the largest absolute effect sizes between ADHD users in r/ADHD ($n = 7{,}758$) and autism users in r/autism ($n = 5{,}706$).

Across 115 LIWC categories, user-level comparison revealed small-to-medium effect sizes (maximum $|d| = 0.76$). ADHD users scored higher on \textsc{time} ($d = 0.76$), \textsc{achieve} ($d = 0.60$), and \textsc{Authentic} ($d = 0.59$), while autism users scored higher on \textsc{socrefs} ($d = -0.56$), \textsc{Social} ($d = -0.52$), and \textsc{Clout} ($d = -0.41$). All confidence intervals are narrow and exclude zero, confirming reliable between-group differences. These results support \textbf{H1}: the two communities maintain distinct and measurable linguistic profiles in their home communities, consistent with prior work on autism-related language patterns \citep{lyons2018}. The full set of 15 significant categories is reported in Appendix~\ref{sec:appendix_full_baseline}.

\begin{table}[t]
\centering
\small
\caption{Baseline linguistic differences (E1, top 10).}
\label{tab:baseline_styles_user}
\begin{tabular}{lccc}
\toprule
\textbf{Category} & \textbf{ADHD} & \textbf{Autism} & \textbf{$d$ [95\% CI]} \\
\midrule
time      & 5.62 & 3.93 & 0.76\,[.72, .80] \\
achieve   & 1.59 & 0.95 & 0.60\,[.56, .64] \\
Authentic & 78.4 & 69.4 & 0.59\,[.55, .63] \\
work      & 2.09 & 1.26 & 0.58\,[.54, .62] \\
socrefs   & 5.05 & 6.68 & $-$0.56\,[.52, .60] \\
substances& 0.46 & 0.05 & 0.55\,[.51, .59] \\
acquire   & 1.29 & 0.83 & 0.53\,[.49, .57] \\
Social    & 8.99 & 11.3 & $-$0.52\,[.48, .56] \\
they      & 0.59 & 0.94 & $-$0.45\,[.41, .49] \\
Lifestyle & 3.19 & 2.35 & 0.45\,[.41, .49] \\
\bottomrule
\end{tabular}
\caption*{\small All FDR-corrected ($q < 0.05$). $+d$: higher in ADHD; $-d$: higher in autism. CIs report absolute $|d|$ bounds.}
\end{table}

\subsection{E2: Cross-Community Accommodation (H2)}

Table~\ref{tab:cross-sectional-accommodation} summarizes the categories with the largest bidirectional shifts. When ADHD users posted in r/autism, they showed decreased \textsc{Authentic} ($d = -0.61$) and \textsc{time} ($d = -0.70$), and increased \textsc{Social} ($d = +0.44$) and \textsc{Clout} ($d = +0.43$). When autism users posted in r/ADHD, the pattern reversed: increased \textsc{Authentic} ($d = +0.33$) and \textsc{time} ($d = +0.43$), decreased \textsc{Social} ($d = -0.47$) and \textsc{Clout} ($d = -0.31$). In each case, the shift direction aligns with the host community's baseline from E1.

As discussed in Section~\ref{sec:operationalize}, the evidence for accommodation is this mirror-image directionality across seven categories. The two largest effects involve \textsc{Authentic} and \textsc{Clout}, which are topic-independent summary variables (Appendix~\ref{sec:appendix_liwc_interp}), providing some reassurance against a purely topical explanation. Effect sizes are small-to-medium ($|d| = 0.31$--$0.70$). These results support \textbf{H2}. Figure~\ref{fig:communities} visualizes the pattern; a quantitative summary is in Appendix~\ref{sec:appendix_barfig}.

\begin{table}[t]
\centering
\small
\caption{Cross-community accommodation (E2).}
\label{tab:cross-sectional-accommodation}
\begin{tabular}{lcc}
\toprule
\textbf{Category} & \textbf{ADHD$\to$Aut.} & \textbf{Aut.$\to$ADHD} \\
 & $d$ [95\% CI] & $d$ [95\% CI] \\
\midrule
Authentic & $-$.61\,[.53,.69] & $+$.33\,[.29,.37] \\
time      & $-$.70\,[.62,.78] & $+$.43\,[.39,.47] \\
Social    & $+$.44\,[.36,.52] & $-$.47\,[.43,.51] \\
Clout     & $+$.43\,[.35,.51] & $-$.31\,[.27,.35] \\
socrefs   & $+$.45\,[.37,.53] & $-$.47\,[.43,.51] \\
i         & $-$.37\,[.29,.45] & $+$.11\,[.07,.15] \\
health    & $-$.25\,[.17,.33] & $+$.44\,[.40,.48] \\
\bottomrule
\end{tabular}
\caption*{\small All FDR-corrected ($q < 0.05$). Every category shifts in opposite directions for the two groups. CIs report absolute $|d|$ bounds.}
\end{table}

\subsection{E3: Exploratory Longitudinal Analysis (RQ)}

We conducted within-user paired comparisons of language use before and after diagnosis disclosure. The full results are reported in Appendix~\ref{sec:appendix_longitudinal}; we summarize the key findings here.

The effects are uniformly small ($|d| \leq 0.22$) with wide confidence intervals. For ADHD users posting in r/autism, only \textsc{health} ($d = -0.22$) and \textsc{prep} ($d = -0.20$) changed significantly, with no changes for \textsc{Authentic} or \textsc{Clout}. For autism users posting in r/ADHD, diagnosis disclosure was associated with decreased \textsc{Authentic} ($d = -0.12$) and increased \textsc{Clout} ($d = +0.11$),notably, in the \emph{opposite} direction from the E2 accommodation pattern for these same categories.

Addressing \textbf{RQ}: diagnosis disclosure is associated with detectable within-user changes, but the effects are weak and should be interpreted as preliminary. Table~\ref{tab:comparison} compares E2 and E3 effect sizes for key categories, showing that accommodation effects are consistently 3--23$\times$ larger than post-diagnosis changes.

\begin{table}[t]
\centering
\small
\caption{E2 (accommodation) vs.\ E3 (post-diagnosis) effect sizes.}
\label{tab:comparison}
\begin{tabular}{lccc}
\toprule
\textbf{Category} & \textbf{Dataset} & \textbf{E2 $d$} & \textbf{E3 $d$} \\
\midrule
\multirow{2}{*}{Authentic} & ADHD$\to$Aut. & $-0.61$ & $-0.09$ \\
 & Aut.$\to$ADHD & $+0.33$ & $-0.12$ \\
\midrule
\multirow{2}{*}{Clout} & ADHD$\to$Aut. & $+0.43$ & $+0.11$ \\
 & Aut.$\to$ADHD & $-0.31$ & $+0.11$ \\
\midrule
\multirow{2}{*}{Social} & ADHD$\to$Aut. & $+0.44$ & $-0.02$ \\
 & Aut.$\to$ADHD & $-0.47$ & $+0.02$ \\
\midrule
\multirow{2}{*}{time} & ADHD$\to$Aut. & $-0.70$ & $-0.15$ \\
 & Aut.$\to$ADHD & $+0.43$ & $-0.00$ \\
\bottomrule
\end{tabular}
\caption*{\small Note directional opposition for Authentic and Clout in the Aut.$\to$ADHD condition.}
\end{table}

\subsection{Summary}

\textbf{H1} is well supported: the two communities maintain distinct baseline linguistic profiles (medium effect sizes, narrow CIs). \textbf{H2} is supported: both groups converge toward the host community's norms when cross-posting, with a systematic mirror-image pattern that is difficult to attribute to topic effects alone (medium effect sizes). The exploratory \textbf{RQ} yields initial but inconclusive evidence: diagnosis disclosure produces small and sometimes directionally opposite effects relative to situational accommodation.

\section{Discussion and Conclusion}

This study applied Communication Accommodation Theory to examine intergroup linguistic dynamics between ADHD and autism communities on Reddit. Our findings can be summarized as follows. First, the two communities maintain distinct and measurable baseline linguistic profiles (H1, medium effect sizes). Second, when users cross community boundaries, their language shifts systematically toward the host community's norms in a mirror-image pattern (H2, medium effect sizes). Third, in an exploratory longitudinal analysis, public diagnosis disclosure produced only small linguistic changes (RQ, $|d| \leq 0.22$), sometimes directionally opposite to cross-community accommodation. While this provides initial evidence that situational and identity-related processes may differ, the small effect sizes and wide confidence intervals warrant cautious interpretation and replication with larger samples.

\paragraph{Practical implications.} Community norms appear to shape language even for visitors from other diagnostic groups, which may be relevant for designing inclusive cross-community spaces. The finding that accommodation is primarily situational suggests that cross-community engagement does not fundamentally alter how users communicate in their home communities.

\paragraph{Implications for clinical understanding.} From a clinical perspective, the distinct linguistic profiles we identify align with recognized communicative differences between ADHD and autism \citep{lyons2018}. The finding that users flexibly adjust these markers when crossing community boundaries adds nuance to how clinicians might interpret language-based signals in online mental health contexts: a diagnostic group's ``characteristic'' language may be more context-dependent than previously assumed. This context-dependence also has implications for NLP systems trained on clinical or social media text. If the lexical features that distinguish diagnostic groups are the same features that shift under social pressure,as recent work on targeted model editing suggests \citep{aithou2026},then classifiers trained on in-community data may be less reliable when applied to cross-community or mixed-audience settings.


\section*{Limitations}

Several limitations should be considered when interpreting these results.

First, our design cannot fully disentangle stylistic accommodation from topical adaptation. When an ADHD user posts in r/autism, they likely discuss different content, which could drive some LIWC shifts. The involvement of topic-independent summary variables (\textsc{Authentic}, \textsc{Clout}) and the within-user design provide partial mitigation, but future work should conduct topic-controlled replications,such as matching posts by topic cluster before computing LIWC differences,and compare cross-posters against non-cross-posters at baseline to rule out selection bias (see Appendix~\ref{sec:appendix_topic} for a proposed methodology).

Second, diagnosis labels were derived from self-disclosed statements in Reddit posts rather than confirmed clinical diagnoses \citep{mindset}. While self-disclosure is an accepted method in social media mental health research, it may introduce misclassification if users speculate about or misrepresent their diagnostic status.

Third, our analysis is limited to two neurodivergent communities on a single platform and may not generalize to other conditions or offline contexts. Fourth, the longitudinal analysis assumes the disclosure timestamp approximates diagnosis timing; delays may blur the pre-/post-boundary. Fifth, user-level aggregation reduces statistical power, and LIWC captures only lexical phenomena,future work should incorporate syntactic, embedding-based, and discourse-level features.

\section*{Ethical Considerations}

This study analyzes publicly available Reddit posts from users who self-disclosed mental health diagnoses. We report no individual-level data, usernames, or verbatim content. All analyses are aggregate-level. We use labels from the Mindset dataset \citep{mindset} and do not assign diagnoses. We acknowledge the risk that analysis of neurodivergent communication could be misused to stigmatize; our intent is to understand intergroup dynamics and inform inclusive platform design.

\bibliography{custom}

@article{pennebaker2015,
  title={The development and psychometric properties of {LIWC2015}},
  author={Pennebaker, James W and Boyd, Ryan L and Jordan, Kayla and Blackburn, Kate},
  journal={University of Texas at Austin},
  year={2015}
}

@inproceedings{danescu2011,
  title={Mark my words! {L}inguistic style accommodation in social media},
  author={Danescu-Niculescu-Mizil, Cristian and Gamon, Michael and Dumais, Susan},
  booktitle={Proceedings of the 20th International Conference on World Wide Web},
  pages={745--754},
  year={2011}
}

@article{giles2015,
  title={Communication accommodation theory},
  author={Giles, Howard and Soliz, Jordan},
  journal={The International Encyclopedia of Interpersonal Communication},
  pages={1--21},
  year={2015},
  publisher={Wiley}
}

@article{giles1973,
  title={Accent mobility: A model and some data},
  author={Giles, Howard},
  journal={Anthropological Linguistics},
  volume={15},
  number={2},
  pages={87--105},
  year={1973}
}

@article{bernhold2022,
  title={Communication accommodation theory},
  author={Bernhold, Quinten S and Giles, Howard},
  journal={The Handbook of Language and Emotion},
  pages={389--407},
  year={2022},
  publisher={Cambridge University Press}
}

@article{tamburrini2015,
  title={Twitter users change word usage according to conversation-loss: A large-scale study},
  author={Tamburrini, Nadine and Cinnirella, Marco and Jansen, Vincent AA and Bryden, John},
  journal={PLOS ONE},
  volume={10},
  number={9},
  pages={e0137149},
  year={2015}
}

@article{kovacs2021,
  title={The role of emotional mimicry in intergroup relations},
  author={Kov{\'a}cs, Bal{\'a}zs and Kleinbaum, Adam M},
  journal={Proceedings of the National Academy of Sciences},
  year={2021}
}

@inproceedings{boghrati2018,
  title={Syntax accommodation in social media conversations},
  author={Boghrati, Reihane and Hoover, Joe and Johnson, Kate M and Garten, Justin and Dehghani, Morteza},
  booktitle={Proceedings of the International Conference on Social Computing},
  year={2018}
}

@article{lyons2018,
  title={Language patterns in autism spectrum disorder},
  author={Lyons, Michael and Aksayli, N Dilara and Brewer, Gayle},
  journal={Journal of Autism and Developmental Disorders},
  volume={48},
  pages={1--12},
  year={2018}
}

@article{guntuku2019,
  title={Studying expressions of loneliness in individuals using {T}witter},
  author={Guntuku, Sharath Chandra and Yaden, David B and Kern, Margaret L and Ungar, Lyle H and Eichstaedt, Johannes C},
  journal={BMJ Open},
  volume={9},
  number={11},
  year={2019}
}

@article{chancellor2020,
  title={Methods in predictive techniques for mental health status on social media: A critical review},
  author={Chancellor, Stevie and De Choudhury, Munmun},
  journal={NPJ Digital Medicine},
  volume={3},
  number={43},
  year={2020}
}

@article{corrigan2015,
  title={Coming out proud to eliminate the stigma of mental illness},
  author={Corrigan, Patrick W and Kosyluk, Kristin A and R{\"u}sch, Nicolas},
  journal={American Journal of Public Health},
  volume={105},
  number={5},
  pages={185--186},
  year={2015}
}

@inproceedings{dearmond2020,
  title={Self-disclosure of mental health on {R}eddit},
  author={De Armond, Serena and others},
  booktitle={Proceedings of the ACM Conference on Computer-Supported Cooperative Work},
  year={2020}
}

@article{rideout2021,
  title={Social media, social life: Teens reveal their experiences},
  author={Rideout, Victoria and Robb, Michael B},
  journal={Common Sense Media},
  year={2021}
}

@article{mindset,
  title={{MindSET}: Advancing Mental Health Benchmarking through Large-Scale Social Media Data},
  author={Mankarious, Saad and Zirikly, Ayah and Wiechmann, Daniel and Kerz, Elma and Kempa, Edward and Qiao, Yu},
  journal={arXiv preprint arXiv:2511.20672},
  year={2025}
}

@article{grootendorst2022,
  title={{BERTopic}: Neural topic modeling with a class-based {TF-IDF} procedure},
  author={Grootendorst, Maarten},
  journal={arXiv preprint arXiv:2203.05794},
  year={2022}
}

@article{walls2006,
  title={Addressing pseudoreplication in longitudinal studies},
  author={Walls, Theodore A and Schafer, Joseph L},
  journal={Models for Intensive Longitudinal Data},
  year={2006},
  publisher={Oxford University Press}
}

@article{benjamini1995,
  title={Controlling the false discovery rate: A practical and powerful approach to multiple testing},
  author={Benjamini, Yoav and Hochberg, Yosef},
  journal={Journal of the Royal Statistical Society: Series B},
  volume={57},
  number={1},
  pages={289--300},
  year={1995}
}

@book{cohen1988,
  title={Statistical Power Analysis for the Behavioral Sciences},
  author={Cohen, Jacob},
  edition={2nd},
  year={1988},
  publisher={Lawrence Erlbaum Associates}
}

@article{boyd2022,
  title={{LIWC-22}: The development and psychometric properties of the latest version of {L}inguistic {I}nquiry and {W}ord {C}ount},
  author={Boyd, Ryan L and Ashokkumar, Ashwini and Seraj, Sarah and Pennebaker, James W},
  journal={Journal of Personality and Social Psychology},
  year={2022}
}

@article{ferrara1991,
  title={Accommodation in therapy},
  author={Ferrara, Kathleen},
  journal={Language and Communication},
  year={1991}
}

@inproceedings{sharma2018,
  title={Mental health support and its relationship to linguistic accommodation in online communities},
  author={Sharma, Eva and De Choudhury, Munmun},
  booktitle={Proceedings of the CHI Conference on Human Factors in Computing Systems},
  year={2018}
}

@article{boorse2019,
  title={Linguistic analysis of narrative production in autism},
  author={Boorse, Jennifer and others},
  journal={Journal of Autism and Developmental Disorders},
  year={2019}
}

@inproceedings{dechoudhury2014,
  title={Mental health discourse on {R}eddit: Self-disclosure, social support, and anonymity},
  author={De Choudhury, Munmun and De, Sushovan},
  booktitle={Proceedings of the International AAAI Conference on Web and Social Media},
  year={2014}
}

@article{depape2025,
  title={Navigating identity and disclosure in emerging adults with {ADHD}},
  author={DePape, Anne-Marie and others},
  journal={Journal of Attention Disorders},
  year={2025}
}

@article{togher2023,
  title={Autism social identification predicts disclosure behaviors},
  author={Togher, Lucy and others},
  journal={Autism Research},
  year={2023}
}

@article{aithou2026,
  title={Parameter-Efficient Token Embedding Editing for Clinical Class-Level Unlearning},
  author={Ait Hou, Iyad and Borad, Shrenik and Sharma, Harsh and Srinivasan, Pooja and Hwa, Rebecca and Zirikly, Aya},
  journal={arXiv preprint arXiv:2603.19302},
  year={2026}
}

\clearpage
\appendix

\section{Quantitative Summary of All Experiments}
\label{sec:appendix_barfig}
\begin{figure*}[ht]
    \centering
    \includegraphics[width=\textwidth]{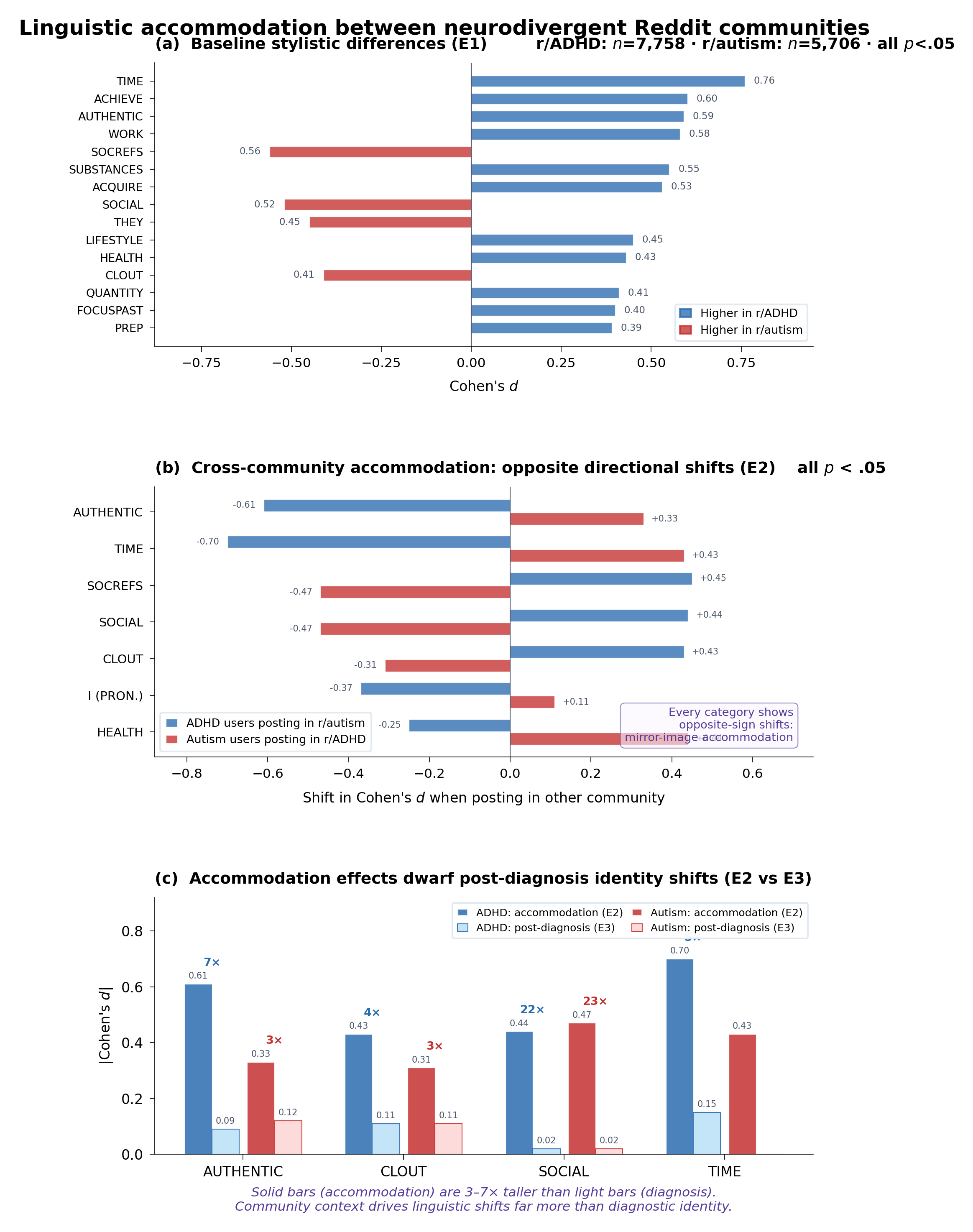}
    \caption{(a)~Baseline differences (E1). (b)~Cross-community accommodation (E2): mirror-imaged shifts across every category. (c)~E2 vs.\ E3: accommodation (solid) is 3--23$\times$ larger than post-diagnosis changes (light).}
    \label{fig:accommodation}
\end{figure*}

Figure~\ref{fig:accommodation} provides a three-panel quantitative summary. Panel~(a) shows baseline stylistic differences (E1), panel~(b) visualizes the mirror-image accommodation shifts (E2), and panel~(c) compares E2 and E3 effect sizes, showing that situational accommodation is 3--23$\times$ stronger than post-diagnosis identity changes.

\section{Full Baseline Differences (E1)}
\label{sec:appendix_full_baseline}
Table~\ref{tab:full_baseline} reports the 15 LIWC categories with the largest absolute effect sizes from E1. These organize into three interpretive clusters.
\emph{Temporal and goal-oriented language}: ADHD users score higher on \textsc{time} ($d{=}0.76$), \textsc{achieve} ($d{=}0.60$), and \textsc{work} ($d{=}0.58$), consistent with ADHD's heightened temporal awareness and goal-directed urgency \citep{lyons2018}.
\emph{Social orientation}: Autism users score higher on \textsc{socrefs} ($d{=}{-}0.56$), \textsc{Social} ($d{=}{-}0.52$), \textsc{they} ($d{=}{-}0.45$), and \textsc{Clout} ($d{=}{-}0.41$), reflecting the community's focus on navigating social relationships.
\emph{Self-presentation}: ADHD users score higher on \textsc{Authentic} ($d{=}0.59$), indicating spontaneous self-expression, while autism users' higher \textsc{Clout} reflects more authoritative language. Both are LIWC summary variables designed to be topic-independent.
\begin{table}[ht]
\centering
\small
\caption{Baseline differences (E1, top 15 by $|d|$).}
\label{tab:full_baseline}
\begin{tabular}{lccc}
\toprule
\textbf{Category} & \textbf{ADHD} & \textbf{Autism} & \textbf{$d$} \\
\midrule
time      & 5.621 & 3.929 & 0.76 \\
achieve   & 1.585 & 0.950 & 0.60 \\
Authentic & 78.41 & 69.37 & 0.59 \\
work      & 2.091 & 1.255 & 0.58 \\
socrefs   & 5.047 & 6.682 & $-$0.56 \\
substances& 0.462 & 0.050 & 0.55 \\
acquire   & 1.293 & 0.833 & 0.53 \\
Social    & 8.987 & 11.33 & $-$0.52 \\
they      & 0.588 & 0.937 & $-$0.45 \\
Lifestyle & 3.185 & 2.350 & 0.45 \\
health    & 3.120 & 2.179 & 0.43 \\
Clout     & 19.96 & 26.62 & $-$0.41 \\
quantity  & 4.306 & 3.544 & 0.41 \\
focuspast & 4.423 & 3.563 & 0.40 \\
prep      & 12.13 & 11.07 & 0.39 \\
\bottomrule
\end{tabular}
\caption*{\small All FDR-corrected ($q < 0.05$). $+d$: ADHD higher; $-d$: autism higher.}
\end{table}

\section{Mirror-Image Accommodation}
\label{sec:appendix_mirror}
Table~\ref{tab:mirror} aligns E1 baselines with E2 shift directions. For every category, features that are high for a group at baseline decrease when that group visits the other community, and vice versa,the defining signature of convergent accommodation.
\begin{table}[ht]
\centering
\small
\caption{E1 baseline direction vs.\ E2 shift direction.}
\label{tab:mirror}
\begin{tabular}{lccc}
\toprule
\textbf{Category} & \textbf{Higher at E1} & \textbf{ADHD$\to$Aut.} & \textbf{Aut.$\to$ADHD} \\
\midrule
Authentic & ADHD & $\downarrow$ & $\uparrow$ \\
time      & ADHD & $\downarrow$ & $\uparrow$ \\
Social    & Autism & $\uparrow$ & $\downarrow$ \\
Clout     & Autism & $\uparrow$ & $\downarrow$ \\
socrefs   & Autism & $\uparrow$ & $\downarrow$ \\
health    & ADHD & $\downarrow$ & $\uparrow$ \\
i         & ADHD & $\downarrow$ & $\uparrow$ \\
\bottomrule
\end{tabular}
\caption*{\small Every shift moves toward the host community's baseline.}
\end{table}

\section{Longitudinal Analysis (E3)}
\label{sec:appendix_longitudinal}
Table~\ref{tab:longitudinal-accommodation} reports all significant within-user changes after diagnosis disclosure. Three observations stand out: (1)~only 2 of 115 categories change for ADHD users and 6 for autism users,disclosure does not broadly reshape linguistic style; (2)~all effects are small ($|d| \leq 0.22$), compared to medium effects in E2 ($|d| = 0.31$--$0.70$); (3)~for autism users in r/ADHD, the post-diagnosis direction for \textsc{Authentic} ($d{=}{-}0.12$) and \textsc{Clout} ($d{=}{+}0.11$) is \emph{opposite} to the E2 accommodation direction ($d{=}{+}0.33$ and $d{=}{-}0.31$), suggesting that identity consolidation and situational accommodation may pull in different directions.
\begin{table*}[ht]
\centering
\small
\caption{Longitudinal changes after diagnosis disclosure (exploratory E3).}
\label{tab:longitudinal-accommodation}
\begin{tabular}{lcccccc}
\toprule
\textbf{Dataset} & \textbf{Category} & \textbf{Pre} & \textbf{Post} & \textbf{Shift} & \textbf{$d$} & \textbf{95\% CI} \\
\midrule
ADHD in Aut. & health & 2.45 & 1.65 & $-0.80$ & $-0.22$ & [$-.36$, $-.08$] \\
 & prep & 11.12 & 10.25 & $-0.87$ & $-0.20$ & [$-.34$, $-.06$] \\
\midrule
Aut.\ in ADHD & Authentic & 77.22 & 74.21 & $-3.00$ & $-0.12$ & [$-.20$, $-.04$] \\
 & Clout & 19.30 & 22.12 & $+2.82$ & $+0.11$ & [$+.03$, $+.19$] \\
 & i & 8.81 & 8.30 & $-0.51$ & $-0.11$ & [$-.19$, $-.03$] \\
 & they & 0.59 & 0.69 & $+0.11$ & $+0.10$ & [$+.02$, $+.18$] \\
 & health & 3.09 & 3.45 & $+0.36$ & $+0.09$ & [$+.01$, $+.17$] \\
 & socrefs & 4.85 & 5.26 & $+0.41$ & $+0.08$ & [$+.00$, $+.16$] \\
\bottomrule
\end{tabular}
\caption*{\small FDR-corrected ($q < 0.05$). All effects small to negligible ($|d| \leq 0.22$).}
\end{table*}

\subsection{Timeline of User Activity}
To contextualize the pre-/post-diagnosis split, Table~\ref{tab:timeline} reports average temporal distances between users' first post, diagnosis disclosure, and final post. ADHD users in r/autism show the shortest pre-diagnosis window ($M = 52$ days), while autism users in r/ADHD show the longest ($M = 304$ days), suggesting that autism-diagnosed users participate in ADHD spaces well before disclosing their diagnosis.

\begin{table}[ht]
\centering
\small
\caption{Average timeline distances (days).}
\label{tab:timeline}
\begin{tabular}{lccc}
\toprule
\textbf{Dataset} & \textbf{First$\to$Dx} & \textbf{Dx$\to$End} & \textbf{First$\to$End} \\
\midrule
ADHD in ADHD   & 124 \scriptsize{(302)} & 568 \scriptsize{(622)} & 692 \scriptsize{(691)} \\
ADHD in Autism  & 52 \scriptsize{(167)}  & 411 \scriptsize{(467)} & 464 \scriptsize{(513)} \\
Autism in ADHD  & 304 \scriptsize{(501)} & 265 \scriptsize{(409)} & 569 \scriptsize{(637)} \\
Autism in Autism & 236 \scriptsize{(386)} & 378 \scriptsize{(458)} & 614 \scriptsize{(576)} \\
\bottomrule
\end{tabular}
\caption*{\small Dx = diagnosis disclosure. Parentheses = SD.}
\end{table}

\section{Proposed Robustness Analyses}
\label{sec:appendix_crossposter}
\label{sec:appendix_topic}

The E2 accommodation results face two key threats to validity: (1)~cross-posters may differ systematically from non-cross-posters at baseline (\emph{selection bias}), and (2)~LIWC shifts may reflect topical rather than stylistic adaptation (\emph{topic confound}). We describe the methodology for addressing each and discuss what the current design already controls for.

\paragraph{Cross-posting rates and asymmetry.}
Of 7,758 ADHD-diagnosed users, 788 (10.2\%) also posted in r/autism. Of 5,706 autism-diagnosed users, 2,305 (40.4\%) also posted in r/ADHD. This ${\sim}4\times$ asymmetry likely reflects ADHD--autism comorbidity rates (estimated 30--80\% clinically), r/ADHD's larger subscriber base ($>$1.8M vs.\ $>$400K), and possible differences in community openness.

\paragraph{Selection bias test.}
We compared in-community LIWC profiles (home subreddit only) of cross-posters vs.\ non-cross-posters within each group: ADHD cross-posters ($n{=}788$) vs.\ non-cross-posters ($n{=}6{,}970$), and autism cross-posters ($n{=}2{,}305$) vs.\ non-cross-posters ($n{=}3{,}401$). If cross-posters already exhibit profiles shifted toward the other community's norms \emph{before visiting it}, the E2 effects may partly reflect who chooses to cross-post rather than how they adapt. Conversely, if cross-posters and non-cross-posters are linguistically indistinguishable in their home community, this strengthens the audience-adaptation interpretation. We note that even if cross-posters differ at baseline, the within-user E2 design (comparing each user's own home vs.\ cross-community posts) partially controls for this, since any baseline difference is constant within a user.

\paragraph{Topic-controlled replication.}
We applied BERTopic \citep{grootendorst2022} with \texttt{all-MiniLM-L6-v2} embeddings, UMAP ($n\_\text{neighbors}{=}15$, $n\_\text{components}{=}5$, $\text{min\_dist}{=}0.0$), and HDBSCAN ($\text{min\_cluster\_size}{=}50$) to all posts across both subreddits. For each topic cluster containing posts from both the home and cross-community conditions, we recomputed the LIWC accommodation comparison. Topic-controlled effect sizes are the weighted average across shared topics. This ensures that any surviving accommodation effects cannot be attributed to differences in discussion content between subreddits. We note that even without this analysis, the involvement of LIWC \emph{summary variables} (Authentic, Clout),which are composite scores designed to be topic-independent,provides partial evidence against a purely topical explanation for the E2 effects.

\paragraph{Why these analyses matter.} If both tests come back clean,cross-posters do not differ at baseline, and effects survive topic matching,the accommodation interpretation is substantially strengthened. If either reveals a confound, it narrows the scope of the claim but does not eliminate it: the mirror-image directionality across seven categories remains difficult to explain by selection or topic alone.

\section{Statistical Methodology and Dataset Details}
\label{sec:appendix_methods}
\label{sec:appendix_ci}
\label{sec:appendix_data}

\paragraph{LIWC-22 categories.} LIWC-22 \citep{boyd2022} produces 115 categories in four tiers: 4 \emph{summary variables} (Analytic, Clout, Authentic, Tone,composite scores relatively robust to topic variation), ${\sim}$25 \emph{linguistic dimensions} (function words, pronouns, verb tense), ${\sim}$50 \emph{psychological processes} (affect, cognition, social processes), and ${\sim}$36 \emph{personal concerns} (work, health, leisure). We tested all 115 per experiment without \emph{a priori} selection, relying on FDR correction to control for multiplicity. The summary variables are particularly important for our accommodation argument because they capture broad stylistic dimensions that are less susceptible to topic confounds than content-specific categories like \textsc{health} or \textsc{work}.

\paragraph{User-level aggregation.} For each user $u$ with posts $\{p_1, \ldots, p_n\}$ in a given condition and LIWC category $c$, we compute a single per-user mean:
\[
\bar{x}_{u,c} = \frac{1}{n} \sum_{i=1}^{n} \text{LIWC}(p_i, c)
\]
Users with fewer than 3 posts in a condition were excluded to ensure stable estimates. This aggregation is critical: without it, prolific users would dominate the analysis, and pseudoreplication would inflate statistical significance.

\paragraph{Statistical testing.} Welch's $t$-test (unequal variances assumed) with effect size $d = (\bar{x}_1 - \bar{x}_2) / s_p$, where $s_p$ is the pooled SD. For E3 paired comparisons, we use the SD of within-user difference scores. All 95\% CIs for $d$ are computed via the noncentral $t$-distribution with noncentrality parameter $\lambda = d \cdot \sqrt{n_1 n_2 / (n_1 + n_2)}$ (independent groups) or $\lambda = d \cdot \sqrt{n}$ (paired). FDR correction follows \citet{benjamini1995} at $q = 0.05$ across $m = 115$ tests per experiment.

\paragraph{Diagnosis timestamps.} For E3, we used the timestamp of each user's first public diagnosis disclosure from the Mindset dataset \citep{mindset}. Users were included only if they had $\geq 3$ posts in both pre- and post-disclosure periods. Median pre-disclosure posting span $\approx$ 14 months; post-disclosure $\approx$ 18 months.

\paragraph{Data pipeline.} Starting from Mindset's user identifiers and diagnostic labels: (1)~re-fetch each user's complete Reddit history via Pushshift (Mindset's original posts are filtered to exclude mental health content), (2)~retain only posts in r/ADHD and r/autism, (3)~verify zero user overlap between diagnostic groups, (4)~remove deleted posts and posts with $<5$ tokens, (5)~run LIWC-22 and aggregate to user-level means. Final corpus spans 2015--2024 (majority 2019--2023).

\section{LIWC Summary Variable Interpretation}
\label{sec:appendix_liwc_interp}
Three LIWC summary variables are central to our analysis. \textsc{Authentic} ($d{=}0.59$, ADHD higher) captures spontaneous, self-disclosing language. \textsc{Clout} ($d{=}{-}0.41$, autism higher) captures confident, authoritative language. \textsc{Social} ($d{=}{-}0.52$, autism higher) captures references to people and interpersonal interactions. Unlike content-specific categories, these are composite scores designed to be relatively topic-independent, making their strong accommodation effects in E2 harder to explain as purely topical artifacts.

It is worth noting that LIWC scores reflect word-level frequency patterns, not communicative intent. A high \textsc{Social} score in the autism community does not mean autistic users are more socially engaged,it means they \emph{talk about} social topics more frequently, which is consistent with the community's focus on navigating social difficulties. Similarly, high \textsc{Authentic} in the ADHD community reflects a linguistic pattern (more first-person pronouns, less formal hedging) rather than a judgment about honesty. These distinctions matter for interpreting the accommodation results: when ADHD users decrease in \textsc{Authentic} while posting in r/autism, this reflects a shift toward more guarded, formal language,not a decrease in sincerity.

A broader limitation of LIWC is that it operates at the word level and cannot capture syntactic structure, pragmatic function, or discourse-level phenomena. For example, LIWC cannot distinguish between a user who asks many questions (a possible sign of deference to the host community) and one who uses question marks in rhetorical statements. Future work could complement LIWC with dependency parsing to measure syntactic complexity, sentence embeddings to compute stylistic similarity in continuous space, or discourse-level features such as hedge frequency and turn-taking patterns. These approaches would test whether the accommodation patterns we observe at the lexical level extend to deeper structural dimensions of language use.

\end{document}